\documentclass[runningheads]{llncs}
\usepackage{graphicx}
\usepackage{comment}
\usepackage{amsmath,amssymb} 
\usepackage{color}

\usepackage{tabularx}
\usepackage{multirow}
\usepackage{ctable}
\usepackage{booktabs}
\newcommand{\ie}{\textit{i}.\textit{e}.}
\newcommand{\eg}{\textit{e}.\textit{g}.}

\newcommand{\gz}[1]{\textcolor{cyan}{#1}}

\begin{document}
\pagestyle{headings}
\mainmatter
\def\ECCVSubNumber{2384}  

\title{Scope Head for Accurate Localization \\ in Object Detection} 

\titlerunning{ECCV-20 submission ID \ECCVSubNumber} 
\authorrunning{ECCV-20 submission ID \ECCVSubNumber} 
\author{Anonymous ECCV submission}
\institute{Paper ID \ECCVSubNumber}

\titlerunning{Scope Head for Accurate Localization in Object Detection}
%
\author{Geng Zhan\inst{1} \and
Dan Xu \inst{2} \and
Guo Lu \inst{4} \and \\
Wei Wu \inst{5}  \and
Chunhua Shen \inst{3} \and
Wanli Ouyang \inst{1}}
\authorrunning{G. Zhan et al.}
%
\institute{The University of Sydney, SenseTime Computer Vision Research Group, Australia \\
\and
University of Oxford, UK \\
\and
The University of Adelaide, Australia\\
\and
Shanghai Jiao Tong University, China 
\and
SenseTime Group Ltd, China}
\maketitle

\begin{abstract}
{Existing anchor-based and anchor-free object detectors in multi-stage or one-stage pipelines have achieved very promising detection performance. However, they still encounter the design difficulty in hand-crafted 2D anchor definition and the learning complexity in 1D direct location regression.  
To tackle these issues, in this paper, we propose a novel object detection framework coined as ScopeNet, which models anchors of each location as a mutually dependent relationship. This approach quantizes the prediction space and employs a coarse-to-fine strategy for localization. It achieves superior flexibility as in the regression based anchor-free methods, while produces 
more precise prediction. Besides, an inherit anchor selection score is learned to indicate the localization quality of the detection result, and we propose to better represent the confidence of a detection box by combining the category-classification score and the anchor-selection score. With our concise and effective design, the proposed ScopeNet achieves state-of-the-art results on COCO.}
\keywords{One-stage Object Detection, Anchors, Anchor-based Detection, Anchor-free Detection, Anchor-dependent Modelling}
\end{abstract}

\section{Introduction}
Object detection is a fundamental task in computer vision. Existing object detection approaches can be mainly categorized into multi-stage~\cite{ren2015faster,he2017mask,cai2018cascade} and one-stage detectors~\cite{lin2017focal,redmon2018yolov3,tian2019fcos}.  Multi-stage detectors usually achieve relatively higher accuracy while more computational steps are basically required. In contrast, one-stage detectors are considered to be simpler, faster, and more flexible in design choices. In this paper we focus on the one-stage object detection problem.

\begin{figure}[t]
    \centering
    \includegraphics[width=0.9\textwidth]{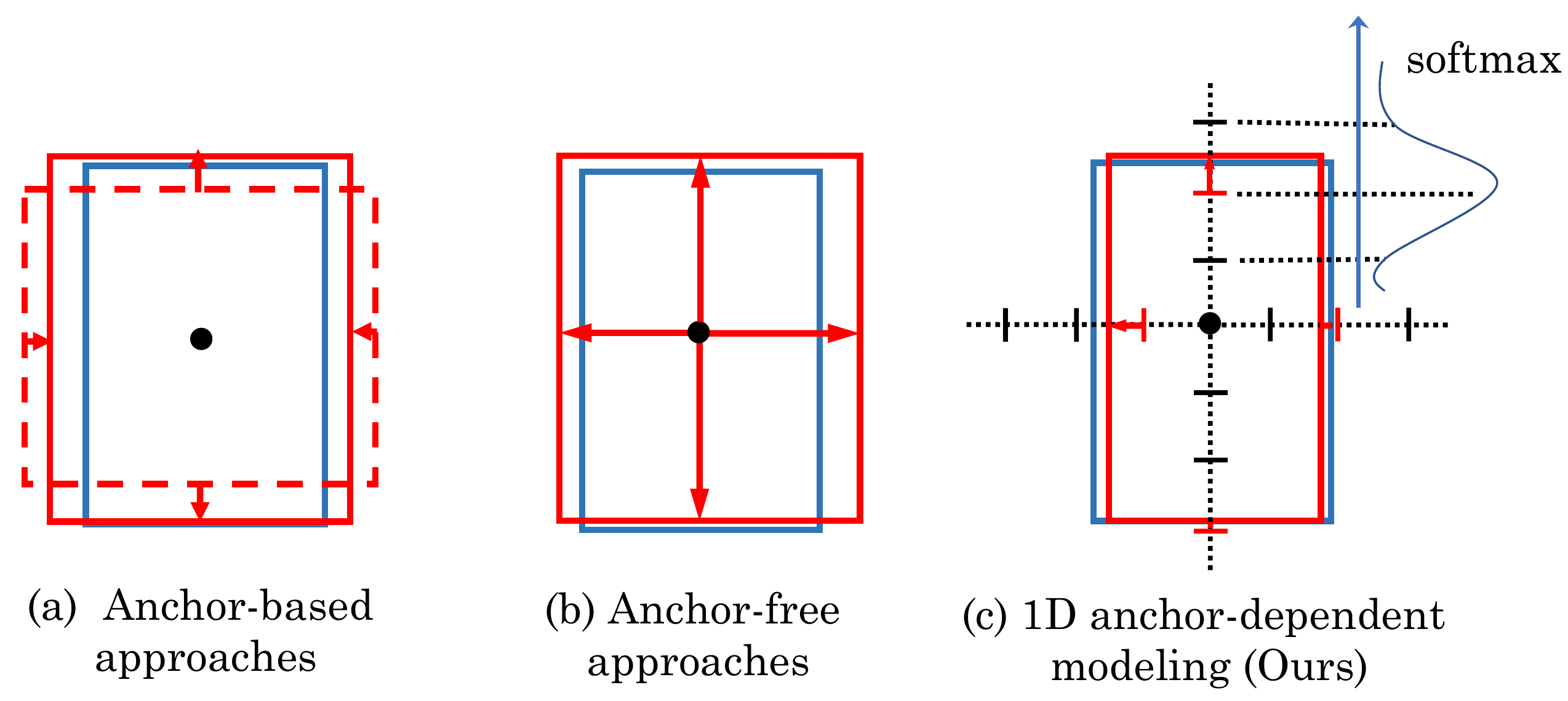}
    \caption{Different approaches for localization prediction. Blue boxes denote ground truth objects. Red boxes in solid line represent the box prediction. Red arrows are regression predictions. Predictions are based on current center locations (black dots). (a) Anchor-based approaches: the dashed box in red denotes one anchor box with possibly high IoU with the target. It first uses classification to select an anchor box that has high IoU with the object, then uses regression to predict a more accurate location. (b) Anchor-free approaches: the four borders of a bounding box are directly regressed. (c) Our 1D anchor-dependent modeling approach: along each of four directions, we first estimate the coarse range of prediction with \textit{softmax} classification, and then apply the corresponding regression network and anchor scale (notch in red) for more fine-grained localization.}
    \label{fig:abc}
\end{figure}

The anchor box definition or generation is a critical component for accurate localization in many state-of-the-art one-stage detectors~\cite{lin2017focal,redmon2018yolov3}. Anchor boxes are generally pre-defined 2D boxes with a set of fixed box shapes, i.e. aspect ratios and scales, at each location (see Fig.~\ref{fig:abc}(a)). The clear benefit of using anchors is that {predicting offsets instead of locations makes it easier for the network to learn~\cite{redmon2017yolo9000}}, and thus facilitates the improvement of the detection accuracy of the detector. However, there are also several drawbacks in commonly used 2D anchor boxes. First, dense prediction over the anchor boxes brings redundancy, as anchors need to be dense enough in order to cover most of the target objects with large variation in box shapes. Then the network has to predict categories and locations for all the anchors. Second, the performance is very sensitive to design choices of anchors. The detection performance remarkably deteriorates with inappropriate anchor designs~\cite{lin2017focal,kong2019foveabox}. Moreover, the representational power of anchors for  objects with varying box shapes is limited. Common one-stage anchor-based detectors~\cite{lin2017focal} typically employ 9 anchors to cover 3 scales and 3 aspect ratios, which clearly makes the detection more challenging to objects with large variation in shapes.



To overcome the above-mentioned disadvantage of 2D anchor boxes, researchers recently proposed anchor-free strategies~\cite{duan2019centernet,zhu2019feature,zhou2019objects,tian2019fcos} via considering direct regression for object localization (see Fig.~\ref{fig:abc}(b)). Through relaxing the box shape constraints on the anchor boxes and learning 1D predictions of four directional offsets (\ie~left, right, top and down) corresponding to the target location, the anchor-free approaches are ideally able to localize objects with arbitrary shape. Despite the simplicity, they achieve comparable or even better performances than anchor-based  methods~\cite{tian2019fcos,kong2019foveabox}. However, these anchor-free methods rely on a single regression network for predicting a precise location in the unbounded space, which 
can be excessively
challenging for the network. 


The analysis above leads to a straightforward question: is it possible to maintain the flexibility and simplicity of anchor-free methods, while simultaneously making the localization more accurate? Since a single regression network 
may not be capable to effectively handle the unbounded prediction space, it is natural for us to consider restricting the range of the regression. Inspired by the anchor-based methods that use distinct regression networks and anchors for objects with different shapes, we also predict different regression values with the corresponding distinct networks and anchors. However, in contrast to common anchor modelling where each anchor prediction is independent to others, we model the anchor existence as a mutually dependent relationship, \ie~there is only one most possible anchor for the object, as depicted in Fig.~\ref{fig:abc}(c). Specifically, we set different 1D anchor scales. For instance, we can divide the {prediction range} into several intervals, each interval corresponding to an anchor scale.
Each anchor scale is responsible for a certain range of locations, and there is one specific regression network for each anchor scale. During the inference, a classification network first produces a coarse prediction and decides which anchor scale and regression network will be used. Then, the corresponding regression network refines the coarse prediction together with the selected anchor scale. This strategy comes with several benefits. First, it balances both the classification and the regression. By quantizing the prediction space, the regression is bounded into a reasonable range. Besides, using~\textit{softmax} for anchor selection 
can make the best-matched anchor compete against other ill-matched ones, thus facilitating more accurate localization. This can effectively rule out the predictions of ill-matched anchors before follow-up possible processing steps (\eg~top-K and NMS), which therefore reduces redundancy and saves computation. Moreover, we enable learning the ranges of the anchor scales, which preserves great flexibility as in existing anchor-free approaches.



To summarize, our main contributions are three-fold:
\begin{itemize}
    \item We model anchors of each location as a mutually dependent relationship and use the \textit{softmax} to make the best-matched anchor compete against ill-matched ones. A coarse-to-fine pipeline is further devised for object localization, which achieves superb flexibility as regression based anchor-free methods, while can clearly reduce the output redundancy and also produce significatly more precise prediction.
    \item We propose an novel and effective strategy to better represent the confidence of a detection box  by combining the category-classification score and the inherit anchor-selection score that indicates the localization quality of the detection result. 
    
    \item We design a consise detection framework with the proposed anchor-dependent modeling and loalization strategies termed as ScopeNet, which establishes state-of-the-art results on COCO without bells and whistles. 
\end{itemize}

\section{Related Work}
\subsection{Anchor-based Detectors}
In deep learning based object detection, inspired from pioneering works such as Faster R-CNN~\cite{ren2015faster} and SSD~\cite{liu2016ssd}, the concept of anchor boxes is widely adopted in later research in the community. For the classic multi-stage detectors~\cite{he2017mask,cai2018cascade,ouyang2017chained}, they select regions of interest (ROIs) which are most likely to contain objects using a Region Proposal Network (RPN) in the first stage, and the anchor box is a basic component in RPN~\cite{ren2015faster}. In the next stages, they first extract deep features from the ROIs~\cite{ren2015faster,he2017mask,gu2018learning}, and two sibling branches are then used for predicting the object class and regressing the object location, respectively. In these pipelines, the anchors are usually ROIs from a previous stage, and the object localization task directly relies on the regression. To improve the localization performance, several recent works consider leveraging classification for localization in further stages. Two representative works are LocNet~\cite{gidaris2016locnet} and Grid R-CNN~\cite{lu2019grid}. LocNet~\cite{gidaris2016locnet} first extracts 1D features along the horizontal and vertical axises, and then performs binary classification to predict the confidence of each position being a bounding box border. It requires several iterative steps to obtain the final results. Grid R-CNN~\cite{lu2019grid} directly predicts the possibility of a border location for each position from its corresponding 2-D feature of each ROI. For single stage detectors, a majority of them are built upon RPN with multi-category classification and regression
~\cite{lin2017focal,liu2016ssd,redmon2018yolov3}. {As far as we know, there is no previous work which models the anchor existence possibility at the same location as a mutually dependent relationship. Besides, existing anchor-based methods basically model anchors as 2D boxes instead of 1D scales as we explore.}

\subsection{Anchor-free Detectors} 
{Based on the localization strategy utilized, anchor-free approaches could be generally divided into classification-based and regression-based ones.}
\subsubsection{Classification based methods.}
 Classification based methods~\cite{law2018cornernet,tychsen2017denet,zhou2019bottom} typically model 2D bounding boxes as sets of points. 
 CornerNet~\cite{law2018cornernet} directly predicts two diagonal corners of the object with classification, and then uses embedding for grouping them into boxes. While DeNet~\cite{tychsen2017denet} also models the box as four corners and predicts them with per-corner classification, it groups the corners by maximizing an overall confidence. ExtremeNet~\cite{zhou2019bottom} predicts one point on each side of the box using semantic meaning, and further considers geometric constraints for the grouping. 
 These classification based methods usually require geometric or semantic relations for grouping corners into boxes in an extra stage, however, the proposed approach directly performs the object classification and localization, and does not require any further grouping or associating processing.
\subsubsection{Regression based methods.}
For two-stage detectors~\cite{wang2019region,he2017mask,he2016deep}, the direct regression is usually performed in region proposal generation and final object localization. 
While among most of recent single-stage detectors~\cite{redmon2016you,duan2019centernet,tian2019fcos}, they model the object localization in an anchor-free fashion. For instance, early works such as YOLO~\cite{redmon2016you} predict the bouding boxes with direct regression of the borders. CenterNet~\cite{duan2019centernet} learns center heatmaps for localizing the object centers instead of corner heatmaps. 
RepPoints~\cite{yang2019reppoints} regresses object boundaries with an iterative dynamic sampling strategy. To develop neat designs, FCOS~\cite{tian2019fcos} and FoveaBox~\cite{kong2019foveabox} recently employ direct regression of a 4D offset vector at each position to represent the four directions of a bounding box for localization. 
All these regression based approaches perform coarse regression on bounding box boarders or offsets without considering bounding the regression range. While in our approach, each regression prediction is only responsible for a certain prediction range, which can effectively achieve more fine-grained and precise localization. Furthermore, our design of using the inherit anchor selection score to better represent the confidence of a detection box is also not investigated in these works.

\section{The Proposed Approach}
\begin{figure*}[t!]
    \centering
    \vspace{20pt}
    \includegraphics[width=\textwidth]{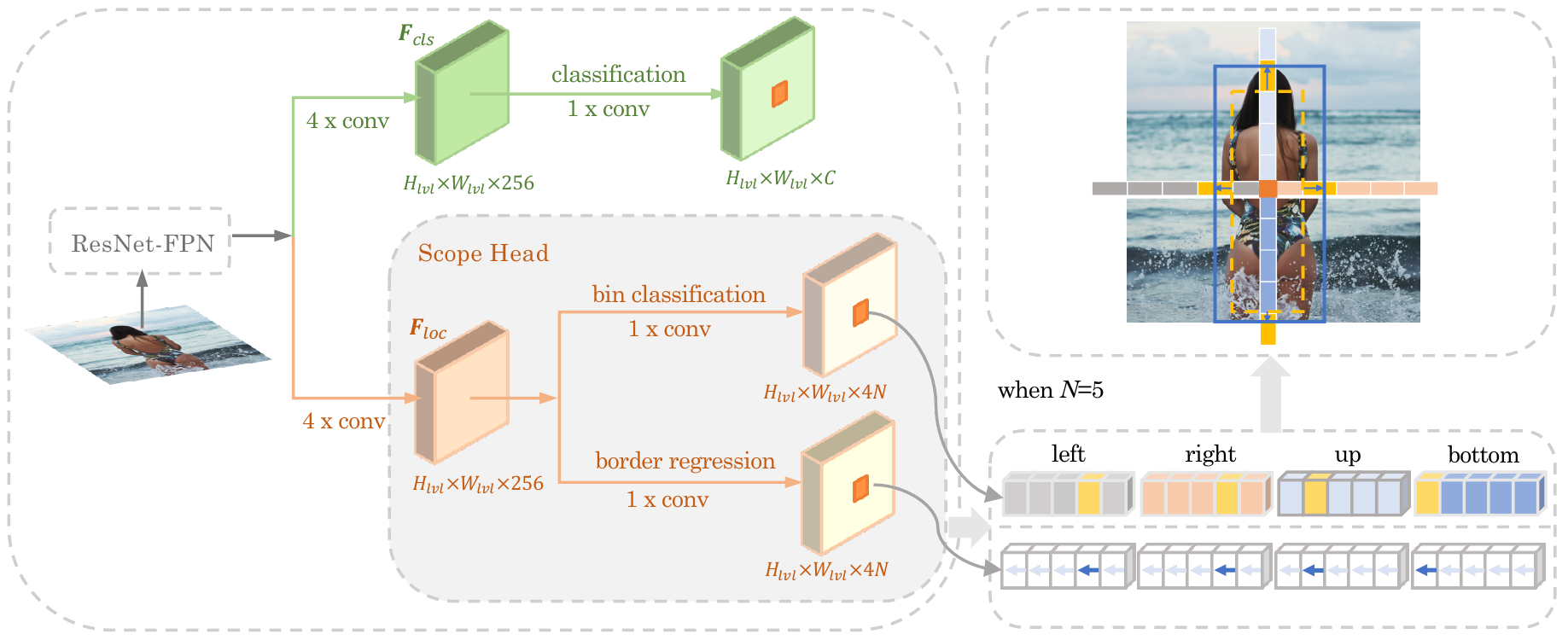}
    \caption{An overview of the proposed detection pipeline ScopeNet. Given an input image, backbone CNN is used for extracting features. Then there are two main branches, object classification branch and object localization branch. The classification branch performs a $C$-class classification for identifying the object category. The object localization branch is implemented by our Scope Head, which is divided into two branches, bin classification, and border regression. The bin classification performs $N$-class classification for anchor selection in four directions. 
The border regression performs border predictions for the anchors selected from the bin classification branch.}
    \label{fig:overview}
\end{figure*}

Our detection pipeline is illustrated in Fig.~\ref{fig:overview}. The backbone network first produces a deep feature map of the input image, which is further used as input of object classification and an object localization network branch, respectively. For each position on the feature map, the classification branch predicts confidence scores of $C$ categories. To predict more accurate localization of objects, the Scope Head detailed in Section~\ref{sec:scopehead} is used at the localization branch.

\subsection{Scope Head}
\label{sec:scopehead}
At the scope head,
we consider $N$ candidate anchors on each of four directions (\ie~left, right, up, and bottom). There are two branches at the scope head, bin classification branch and border regression branch. The bin classification branch is designed to perform a $N$-class classification for learning anchor selection. For these $N$ candidate anchors, the border regression branch performs regression for localizing borders along each direction. Specifically, for each direction, the bin classification branch selects an anchor with the highest score and decodes its corresponding regression prediction to obtain the boundary position of the object in each direction. Finally, the object bounding-box is determined by gathering boundary positions along with four directions.
Note that in our approach, candidate anchors are learnable parameters.

\noindent\textbf{Separate 1D representations.}
\label{subsec:sep_1d_rep}
In most works, an anchor is a 2D representation, \eg~scale, and aspect ratio. In this paper, an anchor is obtained from four separate 1D representations corresponding to the top, bottom, left, and right borders of the anchor. Therefore, the bounding box regression targets are formulated as distances to four boundaries from the position of the current feature point. With such design, it is very flexible to generate anchors with different scales in the 1D space rather than bounding boxes with various shapes in the 2-D space. This naturally brings an inherent advantage, \ie~much higher degree of freedom for the generation of the anchor boxes. For instance, given $N$ anchors for each direction, it could produce anchor boxes with potentially $N^4$ shapes. Therefore, compared with the traditional 2D anchor generation, the 1D anchor representation is more capable of handling various aspect ratios of the objects.

\noindent\textbf{Learnable 1D anchors.} 
In this work, instead of manually designing anchors, we aim at generating learnable anchors that are data-dependent. 
In the 1D anchor representation, there are four directions. We denote a direction by $x$, where $x\in \{l, r, t, b\}$. On each direction $x$, we divide the distance into $N$ bins, where the $n$th bin represents an interval $\mathcal{I}_n=[b_n\  b_{n+1})$, $n=1, \ldots, N$. Each bin has a corresponding anchor. Following the formulation of the target normalization in~\cite{ren2015faster}, we write the boundary prediction for the $n$th bin as follows:
\begin{equation}
\label{eq:anchor_formulation}
    d_{x, n}=A_{n} \cdot e^{t_{x, n}} = 2^{a_{n}} \cdot e^{t_{x, n}},
\end{equation}
where $d_{x, n}$ is the predicted boundary for direction $x$; $A_n$ is the anchor scale, and $t_{x, n}$ is the raw prediction from the border regression network branch. In actual design, the network learns $a_{n}$, where $a_{n}$ is a learnable parameter for the $n$-th anchor, and we use $A_{n}=2^{a_{n}}$. This is because $A_{n}$ might be large (consider existing anchor design, where the maximum value of $A_{n}$ could be $\approx 10^3$). To avoid potential eccentric large value that might make the learning unstable, we predict $a_{n}$ which is in a more reasonable range for network prediction. In previous works, $a_{n}$ is manually designed by assuming a fixed anchor scale and aspect ratio. In our design, $a_{n}$ is learned to adjust the anchor.

\begin{figure}
    \centering
    \includegraphics{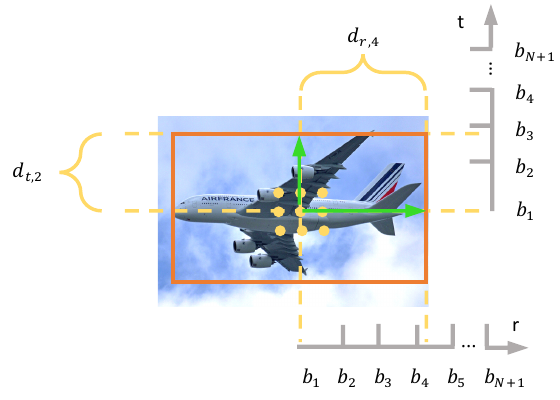}
    \caption{Illustration of the Box parameterization and bin assignment.
    The yellow circles and orange box denote points denote the samples of interest for classification and the target bounding box, respectively. The green vectors that point to the top and left boundaries represent predicted boundary targets $d_{t, 2}$ and $d_{r, 4}$ for the current center position. The formulation for the right and bottom boundaries are analogous to these two while not depicted for simplicity.}
    \label{fig:assign_label}
\end{figure}

\noindent\textbf{Learning anchor selection.}
The choice of $n$ for anchor $A_{n}$ is critical for accurate regression. Only the selected anchor will be used for the regression in Eqn.~(\ref{eq:anchor_formulation}).
Unlike 2D anchor assignment methods where Intersection-over-Union (IoU) metric is adopted, we assign the anchor $A_{n}$ for the boundary $d_x$ by determining which bin the target falls in, as illustrated in Fig.~\ref{fig:assign_label}.
Learning anchor selection aims to select an optimal bin for the target object.
A straightforward method is to use the fully connected layer as the classifier and uses simple  Softmax function to predict probabilities of the target belonging to different bins.

However, different from the classic object classification where the boundary of each category is clear, our bin classification task is ambiguous. 
For example, for the right border close to the boundary of two bins, the network tends to output two high confidence scores for both bins. From our experimental observation, such samples would dominate the loss gradient when the network is trained well on other samples with less ambiguity, which thus degrades the learning performance. To tackle this issue, we propose a strategy which smooths the probability distribution from the Softmax function and down-weighting the loss of such samples as follows: 
\begin{equation}
    p_{x, n}=\dfrac{e^{{s_{x, n}}/{\sigma^2_x}}}{\sum_{m=1}^{N} e^{{s_{x,m}}/{\sigma^2_x}}},
    \label{eq:pxn}
\end{equation}
where $\sigma^2$ is temperature \cite{kaelbling1996reinforcement} indicating the confidence of decision, $p_{x, n}$ is a normalized probability for the direction $x$ within the $n$-th bin and $s_{x,n}$ is the classification score. 
For classification of $N$ classes, the network produces an extra output $s_{x, N+1}$ for predicting $\sigma^2_x$. Therefore, $s_{x, N+1}$ is dependent on the training sample. We set $\sigma^2_x=e^{s_{x, N+1}}$ which gaurantees $\sigma^2$ is positive.

\noindent\textbf{Localization guided detection score.}
To further improve the accuracy of object detection, we employ a localization guided detection score. Most previous works only rely on the classification results to evaluate the qualities of box predictions. 
However, a detection result of high quality not only means category recognized but also requires accurate object localization. Therefore, it is biased to represent the quality of a prediction with merely a classification score.
To address this problem, we use the product of localization confidence and classification score as the score of an object for NMS, which is formulated as
\begin{equation}
\begin{split}
    p_{box}&=p_{cls} \cdot p_{loc}, \\
    p_{loc}&=\dfrac{1}{4}(\tilde{p}_l + \tilde{p}_t + \tilde{p}_r + \tilde{p}_b),
\end{split}
\end{equation}
where $\tilde{p}_x=\underset{n}{\max}(p_{x, n})$, $x \in \{l, r, t, b\}$ denotes the maximum probability of bin classification for border $x$. $p_{x, n}$ is defined in Eqn. (\ref{eq:pxn}). $p_{cls}, p_{loc}$ are classification and localization score for object.

\subsection{Scope Net Details}

  We adopt the Feature Pyramid Network~\cite{lin2017feature} as our baseline network structure. Four stacked convolution layers are used to extract features $\mathbf{F}_{cls}$ and features $\mathbf{F}_{loc}$ for the classification network branch $\mathrm{Net}_{cls}$ and the localization network branch $\mathrm{Net}_{loc}$, as in Fig.~\ref{fig:overview}.
  
  \noindent\textbf{The classification network branch.} It uses the features $\mathbf{F}_{cls}$ for predicting object category scores. For each feature location, $\mathrm{Net}_{cls}$ only predicts one set of multi-class probability scores. While for existing detectors using 2D anchors, due to the existence of multiple anchors per location, multiple sets of scores need to be predicted.
 
   \noindent\textbf{The localization network branch.} $\mathrm{Net}_{loc}$, \ie~the proposed Scope Head Network, takes $\mathbf{F}_{loc}$ as input and produces localization regression scores for each bin as well as the confidence scores for anchor classification, which are used together for the border prediction.

\noindent\textbf{Label assignment for different prediction heads.}
An important step for network training is to assign samples to different FPN levels.
Existing anchor box based approaches assign anchors of different sizes to feature maps of corresponding sizes. In each FPN level, anchors are set as positive samples if the IoU with any ground truth bounding boxes is above a pre-defined threshold. 
While in our approach, each FPN level is responsible for different ranges of regression. We consider a feature point as a positive sample for the $l$-th feature level of FPN based on two factors: (i) the point has to be within a distance to the center of a ground-truth object bounding box, denoted as $r_l$, and (ii) the maximum value of its four location prediction along the four directions, should lie in a reasonable regression range of the $l$-th FPN level. In our experiments, we use $[0, 64)$, $[64, 128)$, $[128, 256)$, $[256, 512)$, $[512, \infty)$ as the regression range for the FPN level from 3 to 7, respectively.  For the classification branch $\mathrm{Net}_{cls}$, we set $r_l$ as $1.5 \times 2^l$, and both the positive and negative samples are used during training. We set $r_l$ as $2 \times 2^l$ for the localization branch $\mathrm{Net}_{loc}$ and only the positive samples contribute in learning $\mathrm{Net}_{loc}$. We use one-hot label for category classification and bin selection prediction. We set the target as directly maximizing the IoU between the decoded bounding box prediction and the target box for border prediction. 

\noindent\textbf{Loss formulation.}
With the ground-truth labels assigned, the overall loss function is defined as:

\begin{equation}
    \mathit{L} = 
    \mathit{L_{cls}} + 1\{c=1\}({\lambda}_{bin}\mathit{L_{bin}} + {\lambda}_{loc}\mathit{L_{loc}}),
\end{equation}

where $1\{\cdot\}$ denotes an indicator function, which returns 1 if $c=1$ \ie~a positive sample, otherwise returns 0. $\mathit{L_{bin}}$ is the bin classification loss and standard Cross Entropy Loss is adopted; $\mathit{L_{cls}}$ is the feature point classification loss and Focal Loss is used as in~\cite{lin2017focal} without parameter tuning; and $\mathit{L_{loc}}$ is the location regression loss where we use the IoU loss following~\cite{yu2016unitbox,tian2019fcos}. Bin classification losses from four directions are averaged as $\dfrac{1}{4} \underset{x}{\sum}\mathit{L}_{loc}^{x}$, where $ x \in \{t, l, r, b\}.$

The loss weights ${\lambda}_{bin}$ and ${\lambda}_{loc}$ are set to 0.5 and 1 respectively, which makes $\mathit{L_{bin}} \approx \mathit{L_{cls}}$ and $\mathit{L_{loc}} \approx 0.5\mathit{L_{cls}}$. We empirically find that this setting stabilize the training process.

\section{Understanding Scope Head Localization}
\label{sec:understanding}

\subsection{{From the View of Anchor-Free Approach}}
{Recently proposed regression based anchor-free methods apply regression only for localization. They are similar to our approach in some aspects. First, we both perform $C$-class classification per location and use similar approaches for choosing positive/negative training samples. Second, we both model the localization predictions as distances to the four borders of the objects. The main difference is whether there is anchor selection. Other anchor-free methods use a regression network to handle the entire prediction space. In fact, such anchor-free methods~\cite{tian2019fcos,kong2019foveabox} belong to a special case of our approach, where $N=1$. While our approach first quantizes the prediction space into several intervals. This bounds each regression in a reasonable range and thus relieves the burden for regression prediction. Besides, the anchor selection score helps to provide information on the localization quality of the predicted box, while existing anchor-free methods lack this information.}

\subsection{{From the Viewpoint of Anchor-based Approach}}
{There are similarities between our approach and anchor-based ones. First, these anchor-based methods first predict the confidence on whether the anchor box matches an object or not. Our approach also selects the anchor scale, which is a 1D anchor that is similar to the 2D anchor box used in existing anchor-based approaches. Second, both anchor-based methods and our approach use regression to refine the location of the anchor with a high score. 
The differences are in three-fold. First, the anchor prediction is dependent in ours but independent in anchor-based approaches. Second, our approach reduces the number of confidence scores required in existing anchor-based methods for anchors.
Assume there are $N$ anchors and $C$ categories. For anchor-based ones, the number of confidence scores is $C \times N$. In our approach, the number is $4\times N + C$. Third, our method can represent anchor boxes more efficiently. In anchor-based approaches, $N$ anchors can only represent $N$ box shapes. While in our approach, $N$ anchors in each direction can represent $N^4$ different shapes of anchor boxes.}

\section{Experiments}

To validate the effectiveness of the proposed ScopeNet detection approach, we conduct experiments on the large scale detection dataset COCO~\cite{lin2014microsoft}. Following~\cite{tian2019fcos,lin2017focal}, we use the~\textit{trainval135k} split as the training set and conduct ablation study on the~\textit{minival} set. {We compare our approach with state-of-the-art methods on the \textit{test-dev} set.}

\subsection{Training Details}
We adopt ResNet~\cite{he2016deep} with FPN~\cite{lin2017feature} as the backbone.
we trained all our models using 8 Nvidia 1080Ti GPUs and a total batch size of 16.
The loss weights for object classification, anchor selection, and border regression are set to be 1, 0.5, and 1, respectively. We use SGD for optimization.

{For ablation study, we train our models for 90K iterations, the $1\times$ schedule. The initial learning rate is 0.01 and is reduced by a factor of 10 at 60K and 80K iteration step. Input images are resized with the shorter side being 800, and the longer side being 1333. No augmentation except horizontal flipping with the probability of 0.5 is adopted.}

{For comparison with state-of-the-art methods, we train our models with the $2\times$ schedule, where we double the iterations to 180K and scale the change points of the learning rate proportionally. The shorter sides of input images range from 640 to 800. We adopt the improvements as mentioned in Table~\ref{tab:tricks}. Other settings are the same as the model with 38.0 mAP in Table~\ref{tab:uncertainty_testing}.}

\subsection{Ablation Study}
\subsubsection{On baseline of direct regression.}
{As discusses in Sec.~\ref{sec:understanding}, our model has fewer differences to anchor-free detectors with regression only. 
Clearly, 
direct regression is a special case of our approach where $N=1$. 
Therefore, we set $N=1$ in our method and use this direct regression model as the baseline.}
As in Table~\ref{tab:component}, this baseline achieves $35.8$ in mAP. 

\subsubsection{On multiple anchor scales and  localization  guided  detection  score.}
Table \ref{tab:component} shows the experimental results on evaluating the new designs in our ScopeNet. 
If our design of multiple anchors is adopted, the mAP is increased from $35.8$ to $37.2$. This validates the effectiveness of using multiple anchors. 
Another advantage is that the anchor-selection score can provide information for the localization quality of the detection box. As in Table~\ref{tab:component}, further utilizing the localization guided detection score can improve $0.8$ on absolute mAP. This information is missing for the direct regression approach, because there is no anchor selection score in such setting.

\begin{table}
    \label{tab:my_label}
    \centering
    \caption{{Improvements upon the baseline. `Direct regression' means the baseline predicting the box with regression only, which corresponds to using one anchor scale with $N=1$. `Multiple anchors' means using multiple anchor scales ($N=5$ here). `Re-score' denotes using  localization confidence for rescoring the detection confidence.}}
    \vspace{3pt}
    \setlength{\tabcolsep}{2pt}
    \begin{tabular}{c|c|c|c c c c c c}
        \toprule
         Direct regression & Multiple anchors & Re-score & $AP$ & $AP_{50}$ & $AP_{75}$ & $AP_{S}$ & $AP_{M}$ & $AP_{L}$ \\
         \midrule
         \checkmark & & & 35.8 & 55.2 & 37.5 & 18.7 & 40.1 & 47.6 \\
         & \checkmark & & 37.2 & 56.0 & 39.4 & 19.6 & 41.2 & 49.6 \\
         & \checkmark & \checkmark & {38.0} & {56.1} & {40.6} & {20.8} & {41.9} & {50.1} \\
         \bottomrule
    \end{tabular}
    \label{tab:component}
\end{table}

\subsubsection{Parameter choices of anchors.}
In this part, we present experimental results on several key factors for designing anchors in our model. We mainly discuss how the prediction range of each anchor should be assigned, and the effect of the number of anchors. 

The prediction range of each anchor and the number of anchors provide a trade-off on classification and regression for localization. When more anchors are used, the network relies more on classification for selecting a more fine-grained range for the following regression. 
Thus, 
the pressure for regression is eased since it focuses on a smaller prediction range. On the contrary, when fewer anchors are adopted, the regression needs to handle a larger range. A good balance that effectively leverages classification and regression could produce a better result.
To this end, we vary the prediction range for each anchor and the number of anchors. The results are shown in Table \ref{tab:grid-search}.

\label{subsec:hype_sensitive}
In general, it is clear that dividing the whole prediction space into several sub-spaces is better when compared with using single regression handling the entire prediction space. Also, it is also observable that our Scope Head is not sensitive to the hyper parameter choices. It indicates that the classification and regression could be better balanced and trained in different settings. While for detectors with independent 2-D anchors, experimental results in \cite{lin2017focal,kong2019foveabox} show that the performance is more sensitive to the design choices of anchors.

\begin{table}[t]
    \centering
    \caption{Effects of different parameter choices of anchors. We try different sub-space sizes and anchor numbers. The sizes are given as $\log_{2}(\cfrac{b_{n+1}}{b_n})$. Scope Head is not sensitive to hyper parameter choices.}
    \vspace{3pt}
    \setlength{\tabcolsep}{5pt}
    \begin{tabular}{c c|c c c c c c}
    \toprule
        size & num  & $AP$ & $AP_{50}$ & $AP_{75}$ & $AP_{S}$ & $AP_{M}$ & $AP_{L}$ \\
    \midrule
        -       & 1 & 35.8 & 55.2 & 37.5 & 18.7 & 40.1 & 47.6 \\
        $0.75$  & 3 & 36.9 & 55.9 & 38.9 & 19.3 & 40.8 & 48.5 \\ 
        $0.75$  & 5 & 37.0 & 55.6 & 39.2 & 19.3 & 41.2 & 48.6 \\ 
        $0.75$  & 7 & 36.9 & 55.4 & 39.4 & 19.7 & 40.6 & 49.1 \\
        $1$     & 3 & 37.1 & 56.0 & 39.4 & 20.1 & 41.1 & 48.5 \\ 
        $1$     & 5 & 37.2 & 56.0 & 39.4 & 19.6 & 41.2 & 49.6 \\ 
        $1$     & 7 & 37.2 & 56.1 & 39.4 & 20.2 & 40.8 & 49.4 \\ 
        $1.25$  & 3 & 37.1 & 56.1 & 39.5 & 19.3 & 40.8 & 49.6 \\ 
        $1.25$  & 5 & 37.0 & 56.0 & 39.3 & 19.5 & 41.0 & 49.6 \\ 
        $1.25$  & 7 & 36.9 & 55.4 & 39.4 & 19.7 & 40.6 & 49.1 \\ %
        \bottomrule
    \end{tabular}
    \label{tab:grid-search}
\end{table}

\subsubsection{Localization guided box score and uncertainty term.} To show the effectiveness of incorporating localization score in box score and the uncertainty term, we provide results in Table \ref{tab:uncertainty_testing}. When no localization score is used, i.e. $p_{box}=p_{cls}$, $37.2$ mAP is achieved. Next, when using localization score, i.e. $p_{box}=p_{cls} \cdot p_{loc}$, a model without uncertainty term $\sigma^2$ achieves $37.7$ in mAP, which is $0.5$ point higher than the previous one. In comparison, our model using uncertainty as in Eqn. (\ref{eq:pxn}) achieves $38.0$ in mAP, which is $0.3$ point higher than the previous model with vanilla Softmax function.

\begin{table}[t]
    \centering
    \caption{Incorporate anchor confidence and uncertainty for estimating bounding box score.}
    \setlength{\tabcolsep}{5pt}
    \begin{tabular}{c c|c c c c c c}
    \toprule
        $p_{loc}$ & $\sigma^2$ & $AP$ & $AP_{50}$ & $AP_{75}$ & $AR_{10}$ & $AR_{100}$ \\
    \midrule
        - & - & 37.2 & 56.0 & 39.4 & 51.7 & 54.8 \\
        \checkmark & - & 37.7 & 55.2 & 40.5 & 52.3 & 55.1 \\
        \checkmark & \checkmark & 38.0 & 56.1 & 40.6 & 52.5 & 55.4 \\
    \bottomrule
    \end{tabular}
    \label{tab:uncertainty_testing}
\end{table}

\subsubsection{With strategies from other methods.} {We show that the proposed method works well with custom strategies that could boost performance and are adopted in other works. Table~\ref{tab:tricks} reports the results of our approach equipped with different custom strategies. This validates the compatability of our approach with other components. }

\begin{table*}[t]
    \centering
    \caption{Improvements on COCO \textit{minival} set with free use of extra components. {$\Delta$ represents the relative gain against the baseline model without using extra components. `Custom FPN' is introduced in FCOS~\cite{tian2019fcos}. `NMS' means changing the NMS threshold from 0.5 to 0.6. `GN' means using Group Normalization~\cite{wu2018group} in the detection head.}}
    \vspace{3pt}
    \setlength{\tabcolsep}{5pt}
    \begin{tabular}{c|c|c|c|c}
    \toprule
          Custom FPN & NMS & GN & $mAP$ & $\Delta$ \\
    \midrule
     & & & 38.0 & - \\
     \checkmark & & & 38.1 & 0.1 \\
     \checkmark & \checkmark & & 38.4 & 0.4\\
     \checkmark & \checkmark & \checkmark & 39.4 & 1.4 \\
    \bottomrule
    \end{tabular}
    \label{tab:tricks}
\end{table*}

\subsubsection{Generalization to input of different sizes.}
Besides less sensitive to hyperparameter choices, as mentioned in Sec. \ref{subsec:hype_sensitive}, our anchor design also generalizes well for inputs with different sizes. In Table \ref{tab:input_size}, we compared our anchor design with vanilla anchors on various input (image and box) sizes.
For our anchor design, we use our ScopeNet with 5 anchors and the uncertainty term. While for vanilla anchor design, we use RetinaNet with 9 anchors in mmdetection without modifications. For both methods, we use the same backbone structure and only change the anchor scheme. It is clear that our anchor modelling consistently outperforms vanilla anchors in terms of handling input of different sizes.

\begin{table}[t]
    \centering
    \caption{Vanilla anchors vs. our anchors on inputs with different sizes.}
    \setlength{\tabcolsep}{3pt}
    \vspace{3pt}
    \begin{tabular}{c|c|c c c c c c}
    \toprule
         Anchor & input size & $AP$ & $AP_{50}$ & $AP_{75}$ & $AP_{S}$ & $AP_{M}$ & $AP_{L}$ \\
    \midrule
         Vanilla & \multirow{2}{*}{400} & 30.6 & 48.6 & 32.4 & 11.0 & 34.2 & 47.5 \\
         Ours   & & 32.5 & 49.8 & 34.2 & 13.5 & 34.8 & 49.6 \\
    \hline
         Vanilla & \multirow{2}{*}{600} & 34.3 & 53.7 & 36.8 & 16.6 & 38.2 & 48.1 \\
         Ours   & & 35.9 & 54.3 & 38.1 & 17.5 & 39.6 & 50.0 \\
    \hline
         Vanilla & \multirow{2}{*}{800} & 35.8 & 55.5 & 38.3 & 20.1 & 39.5 & 47.7 \\
         Ours   & & 38.0 & 56.1 & 40.6 & 20.8 & 41.9 & 50.1 \\
    \bottomrule
    \end{tabular}
    \label{tab:input_size}
\end{table}

\subsection{Comparison with State-of-the-art Detectors}
{We compare ScopeNet with other state-of-the-art object detectors on the \textit{test-dev} split of COCO dataset. As shown in Table~\ref{tab:sota}, our method surpasses both anchor-based RetinaNet and anchor-free FCOS by a clear margin with the same backbone.}
\begin{table*}[!t]
    \centering
    \caption{COCO \textit{test-dev} results for ScopeNet and other state-of-the-art approaches. 'Anchor' means anchor boxes are utilized.}
    \setlength{\tabcolsep}{1.5pt}
    \vspace{3pt}
    \begin{tabular}{l|c|c|c c c c c c}
    \toprule
         Method & Anchor & Backbone & $AP$ & $AP_{50}$ & $AP_{75}$ & $AP_{S}$ & $AP_{M}$ & $AP_{L}$ \\
         \midrule
         Faster R-CNN~\cite{lin2017focal} & \checkmark & ResNet-101-FPN & 36.2 & 59.1 & 39.0 & 18.2 & 39.0 & 48.2 \\
         Mask R-CNN~\cite{he2017mask} & \checkmark & ResNet-101-FPN & 38.2 & 60.3 & 41.7 & 20.1 & 41.1 & 50.2 \\
         Cascade R-CNN~\cite{cai2018cascade} & \checkmark & ResNet-101-FPN & 42.8 & 62.1 & 46.3 & 23.7 & 45.5 & 55.2 \\
         \hline
         YOLO v3~\cite{redmon2018yolov3} & \checkmark & Darknet-53 & 33.0 & 57.9 & 34.4 & 18.3 & 35.4 & 41.9 \\ 
         CornerNet~\cite{law2018cornernet} & & Hourglass-104 & 40.6 & 56.4 & 43.2 & 19.1 & 42.8 & 54.3 \\
         CenterNet~\cite{duan2019centernet} & & Hourglass-104 & 44.9 & 62.4 & 48.1 & 25.6 & 47.4 & 57.4 \\
         ExtremeNet~\cite{zhou2019bottom} & & Hourglass-104 & 40.2 & 55.5 & 43.2 & 20.4 & 43.2 & 53.1 \\
         TridentNet~\cite{li2019scale} & \checkmark & ResNet-101 & 42.7 & 63.6 & 46.5 & 23.9 & 46.6 & 56.6 \\
         RepPoints~\cite{yang2019reppoints} & & ResNet-101-FPN & 41.0 & 62.9 & 44.3 & 23.6 & 44.1 & 51.7 \\
         RetinaNet~\cite{lin2017focal} & \checkmark & ResNet-101-FPN & 39.1 & 59.1 & 42.3 & 21.8 & 42.7 & 50.2 \\ 
         FSAF~\cite{zhu2019feature} & \checkmark & ResNet-101-FPN & 40.9 & 61.5 & 44.0 & 24.0 & 44.2 & 51.3 \\
         Fovea~\cite{kong2019foveabox} & & ResNet-101-FPN & 40.6 & 60.1 & 43.5 & 23.3 & 45.2 & 54.5 \\ 
         FCOS~\cite{tian2019fcos} & & ResNet-101-FPN & 41.5 & 60.7 & 45.0 & 24.4 & 44.8 & 51.6 \\ 
         \hline
         ScopeNet (ours) & - & ResNet-101-FPN & 43.4 & 61.2 & 47.8 & 26.0 & 46.8 & 53.8 \\
         \bottomrule
    \end{tabular}
    \label{tab:sota}
\end{table*}
In detail, our model achieves comparable results on $AP_{50}$. While in terms of $AP_{75}$, our model clearly surpasses all the other methods (except CenterNet with quite different settings~\cite{duan2019centernet}). This fully demonstates that our model performs better at accurate localization. 


\subsection{{Qualitive Results}}

\begin{figure}[t]
    \centering
    \includegraphics[width=0.7\textwidth]{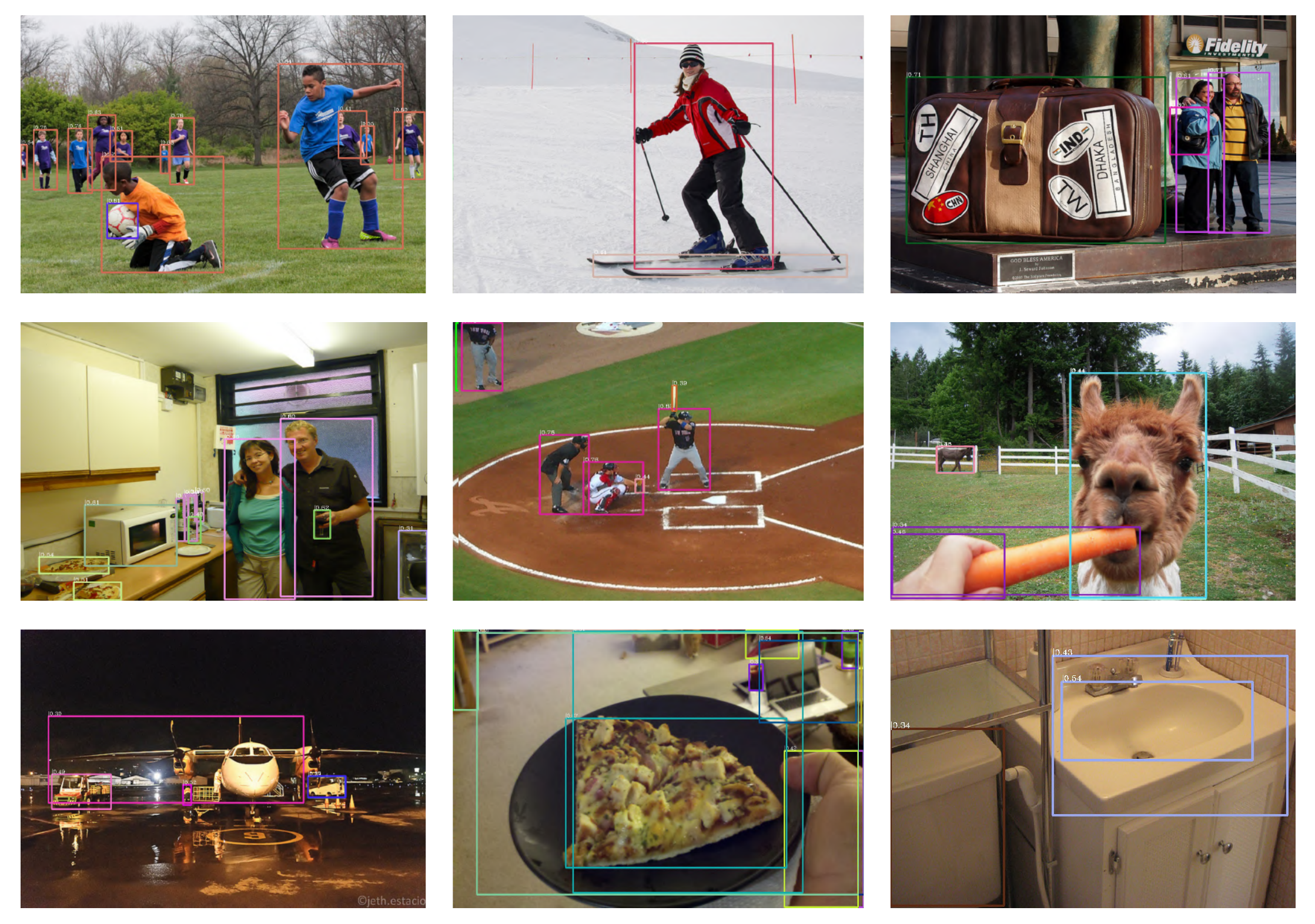}
    \caption{Visualization of detection results. Examples shown in the $1st$ and $2nd$ rows are successful detection samples. Examples in the $3rd$ row contain failure detection cases. For readability, we only draw bounding boxes with prediction scores while not showing class labels on the bounding boxes.}
    \label{fig:qualitive}
\end{figure}

{We provide visualization results of our approach in Fig.~\ref{fig:qualitive}. We use the model that achieves $38.0$ on $mAP$. We select images contain objects with different aspect ratios and crowded scenes. From those images, we could observe that our approach is able to localize objects accurately if the objects have a reasonable size and can be successfully recognized by the classification network.}

\section{Conclusion}
We have presented the proposed framework ScopeNet for object detection. It models anchors of each location as a mutually dependent relationship and considers a coarse-to-fine pipeline for object localization. The proposed approach achieves a great flexibility as in other regression based anchor-free methods, while it also clearly reduces the output redundancy and produces 
better prediction. Moreover, ScopeNet proposes a novel scheme via combining the category-classification score and the inherit anchor selection score that indicates the localization quality of the detection result, which has been shown to be very effective to represent the confidence of a detection box. Extensive experiments demonstrate that the proposed ScopeNet could clearly 
achieve state-of-ther-art results
on the COCO dataset.

\section{Acknowledgement}
We would like to thank Zhi Tian for 
constructive 
discussions.

\clearpage
%
%
\bibliographystyle{splncs04}
\bibliography{eccv2020release}
\end{document}